\documentclass{article}
\usepackage{spconf,amsmath,graphicx,array,booktabs,xcolor,hyperref,tabularx,comment}
\usepackage[absolute,showboxes]{textpos}

\TPMargin{5pt}

\newcommand{\copyrightstatement}{
    \begin{textblock}{0.84}(0.08,0.93)    
         \noindent
         \footnotesize
         \copyright 2022 IEEE.  Personal use of this material is permitted.  Permission from IEEE must be obtained for all other uses, in any current or future media, including reprinting/republishing this material for advertising or promotional purposes, creating new collective works, for resale or redistribution to servers or lists, or reuse of any copyrighted component of this work in other works.
    \end{textblock}
}

\newcommand\BUT{$^2$} 
\newcommand\USAAR{$^1$} 

\title{Call-sign recognition and understanding for noisy air-traffic transcripts using surveillance information}
\name{Alexander Blatt\USAAR, Martin Kocour\BUT, Karel Vesel\'{y}\BUT, Igor Szöke\BUT and Dietrich Klakow\USAAR }
\address{\USAAR Saarland University, Saarland Informatics Campus, Germany \\
\BUT Brno University of Technology, Faculty of Information Technology, Speech@FIT, Czechia
\thanks{The work was supported by European Union’s Horizon 2020 project No. 864702 - ATCO2.}
}
\begin{document}
\copyrightstatement
%
\maketitle
\begin{abstract}
Air traffic control (ATC) relies on communication via speech between pilot and air-traffic controller (ATCO). The call-sign, as unique identifier for each flight, is used to address a specific pilot by the ATCO. Extracting the call-sign from the communication is a challenge because of the noisy ATC voice channel and the additional noise introduced by the receiver. A low signal-to-noise ratio (SNR) in the speech leads to high word error rate (WER) transcripts. We propose a new call-sign recognition and understanding (CRU) system that addresses this issue. The recognizer is trained to identify call-signs in noisy ATC transcripts and convert them into the standard  International Civil Aviation Organization (ICAO) format. By incorporating surveillance information, we can multiply the call-sign accuracy (CSA) up to a factor of four. The introduced data augmentation adds additional performance on high WER transcripts and allows the adaptation of the model to unseen airspaces.
\end{abstract}

\section{Introduction}
\label{sec:intro}

The classical communication between air-traffic controllers (ATCOs) and pilots is voice-based \cite{Eskilsson2020}. This form of communication has the drawback, that one ACTO talks to multiple pilots over a single frequency. The rising traffic in the last years raised the number of pilots tuned in the same frequency. This increases the chance, that two pilots speak simultaneously. To avoid responses from multiple pilots, the ATCO addresses the target airplane by its call-sign. A call-sign is an unique identifier that is assigned to each airplane (e.g. \texttt{DLH83K}). New systems like controller–pilot data link communications (CPDLC), which use text based communication, reduce the load on the voice communication channels \cite{Eskilsson2020}. Projects like AcListant  and MALORCA\footnote{MALORCA Homepage: \url{https://www.malorca-project.de/}} aim to support the ATCO by speech recognition systems \cite{Kleinert2018,Srinivasamurthy2018a}. The problem with developing such systems, is the lack of training data in the ATC domain. Although there exist some datasets \cite{Zuluaga-Gomez2020b}, there is missing a database covering a multitude of locations and containing speech, transcripts and meta information like call-signs and commands. This work is part of the ATCO2 project\footnote{ATCO2 Homepage: \url{https://www.atco2.org/}}, which aims among others to build up such a database.\par 
In this work we are investigating the benefit of including context information for call-sign recognition and understanding. The context information in form of a list of surveillance call-signs is used as additional input for our models. The models recognize the call-sign in an ATC transcript and convert it to the standard ICAO format. For the training of our models, we introduce a data augmentation method, that is adjustable to the target airspace. We can show, that the models trained on the augmented data predict the target call-signs with high accuracy. We also find that the models incorporating surveillance information are superior and show a high resistance to ASR noise and surveillance data variations.

\section{Related Work}
Various works have already investigated context incorporation in the ASR \cite{Shore, Schmidt,Oualil}, which marks the prior step in the ATC speech processing pipeline. Two other works of the ATCO2 project \cite{KocourMartin2019,Nigmatulina2021} show that the combination of HCLG and lattice boosting using Kaldi \cite{Povey2011}, reduces the ATC-ASR errors, especially for the call-signs. We build on top of these works by extracting the (erroneous) call-signs from the ASR transcripts and map them to the standardized ICAO format.
 
In named-entity recognition (NER) the call-sign sequence is identified in the input (Recognition), therefore it is related to our method, which additionally converts the call-sign to the target ICAO format (Understanding). NER for call-signs as single entity of interest is also part of the Airbus challenge \cite{Pellegrini2018}. One of the top three contestants uses a Bi-LSTM-CRF architecture \cite{Gupta2019a} for the call-sign recognition, reaching an F1 score of 80.17 on the leader board. Newer pretrained transformer based models like BERT typically outperform recurrent architectures like LSTMs in natural language processing (NLP) and natural language understanding (NLU) tasks \cite{Devlin2019}.


\section{Data}\label{sec:data}

\autoref{tab:datasets} contains the datasets, that are used for training and testing. The Malorca dataset \cite{Kleinert2018,Srinivasamurthy2018a} consists of transcripts of ATCO speech from the Vienna airport together with surveillance call-signs for each transcript. The LiveATC dataset contains transcripts of ATC speech from Zurich Airport (LSZH) and  Dublin Airport (EIDW) with some samples from Hartsfield–Jackson Atlanta International Airport (KATL). The speech data is collected during the ATCO2 project from LiveATC\footnote{LiveATC Homepage: \url{https://www.liveatc.net/}}, which provides live ATC radio feeds.
\begin{table}[ht]
\centering
\caption{Overview of the datasets. The last column marks the WER of the different versions of the same dataset.}
\begin{tabular}[t]{lcc}
\toprule
Datasets & Samples & WER Variants  \\
\midrule
LiveATC  & 500 &  0$|$28.4 (h)$|$28.9 (l)$|$33.1 (b)\\
Malorca  & 1130 & 0$|$6.42 (h)$|$7.27 (l)$|$8.47 (b)\\
Airbus  & 60000& 0$|$7.00$|$30.0\\
\bottomrule
\end{tabular}
\label{tab:datasets}
\end{table}%
The Malorca and LiveATC transcripts are generated by three different ASR methods (baseline (b), lattice-boosting (l) and HCLG-lattice boosting (h)) \cite{KocourMartin2019} and by human transcription for the ground-truth data (WER 0). All transcripts are manually annotated with the correct ICAO call-sign. The generation of the augmented Airbus datatset out of the Airbus development dataset \cite{Delpech2019} is described in \autoref{sec:augmentation}.\par A sample of the datasets consists out of the  transcript (\texttt{lufthansa eight three kilo descend three thousand feet}), the corresponding target ICAO call-sign (\texttt{DLH83K}) and the surveillance call-signs (\texttt{AIF44T}, \texttt{DLH83K}, \texttt{MAN47N}, ...). The surveillance data is drawn from the OpenSky Network\footnote{OSN Homepage: \url{https://opensky-network.org/}} (OSN) database \cite{Schafer2014}. We isolate the call-signs from the surveillance–broadcast (ADS-B) data fetched for each transcript and use them as context information. On average a sample contains 26 (Malorca), respectively 30 (LiveATC) surveillance call-signs.\par  
The call-signs start generally with an airline identifier\footnote{A list of identifiers can be found here: \url{https://en.wikipedia.org/wiki/List_of_airline_codes}} (\texttt{lufthansa} $\leftrightarrow$ \texttt{DLH}) followed by an alphanumeric call-sign number (\texttt{eight three kilo} $\leftrightarrow$ \texttt{83K}). The call-sign number in the transcript is converted to its ICAO equivalent by the use of the NATO phonetic alphabet\footnote{NATO phonetic alphabet: \url{https://en.wikipedia.org/wiki/NATO_phonetic_alphabet}}.


\section{Data Augmentation}\label{sec:augmentation}
Each airspace has distinct characteristics like the occurrence of regional airlines. Noise levels of the voice channel can also vary, resulting in different WERs of the transcripts. Ideally, a CRU system could be fine-tuned to each new airspace by training it on a database for this region. In reality, there exist only a handful of ATC databases \cite{Zuluaga-Gomez2020b}. But not all of them contain labeled call-signs. A timestamp, as well as the location of the recordings are also missing in most cases, which makes it impossible to retrieve the corresponding surveillance information from the OSN database.\par
\begin{figure}[htb]
\begin{minipage}[b]{1.0\linewidth}  
  \centering
  \centerline{\includegraphics[width=7cm]{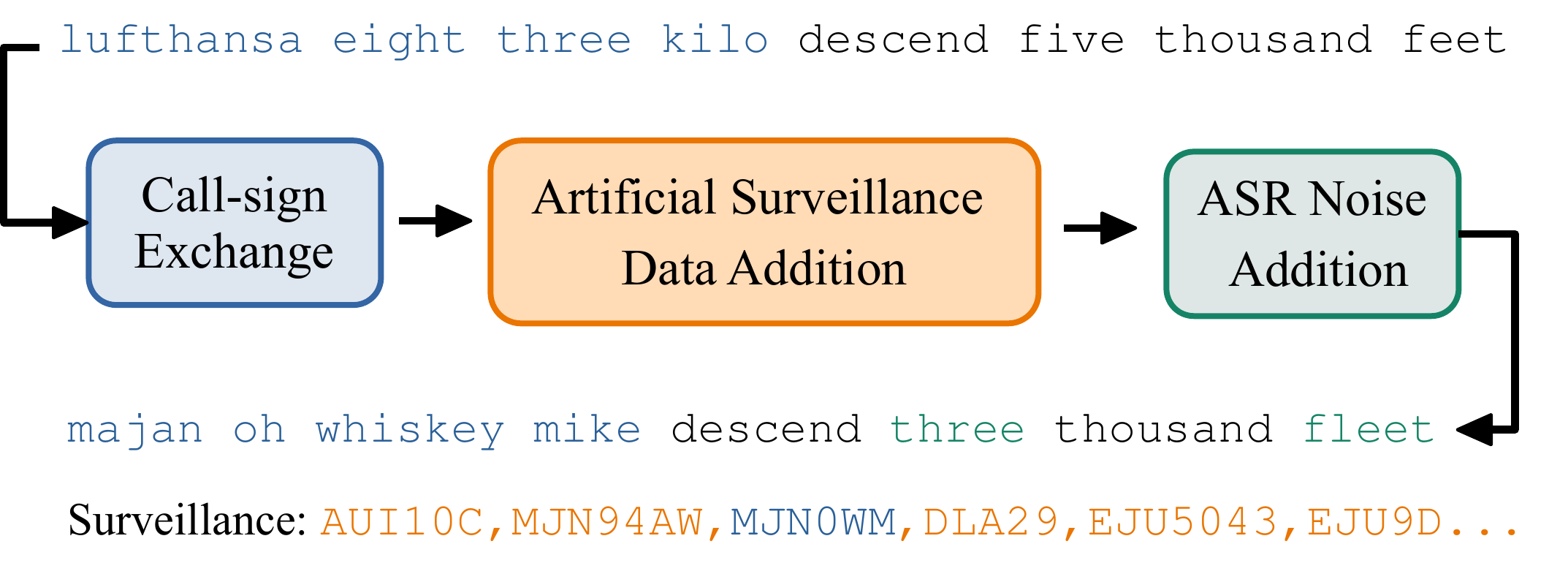}}
\end{minipage}
\caption{Scheme of the data augmentation pipeline.}
\label{fig:augmentation}
\end{figure}
To overcome this issue, we propose a data augmentation pipeline which is shown in \autoref{fig:augmentation}. The basis for the pipeline is the Airbus training dataset \cite{Delpech2019}, which contains roughly 28.000 transcripts with labeled call-signs. In the first step of the augmentation, the call-sign (\texttt{lufthansa eight three kilo}) is cut out of the transcript and replaced with an artificially generated call-sign (\texttt{majan oh whiskey mike}).\par 
The rule-based data augmentation also includes real-life variations from the standard format. This includes missing identifiers, shortened call-sign numbers and the usage of different identifier formats. Transcript equivalents of \texttt{DLH72K} are for example \texttt{lufthansa seven two kilo}, \texttt{seven two kilo}, \texttt{lufthansa}, \texttt{lufthansa seventy-two kilo} and \texttt{dlh seven two kilo}.\par
In the next step, surveillance call-signs are added with the same parameters (number of call-signs with the same identifier, number of total call-signs, surveillance length) as real surveillance. To match the noise level of the test datasets  (Malorca and LiveATC), simulated ASR noise (noisy distribution extracted from noisy ASR output) is introduced in the last step for the two noisy datasets (WER 7.0 and WER 30.0) but not for clean dataset (WER 0). 

\section{Context Integration}
\begin{figure}[htb]
\begin{minipage}[b]{1.0\linewidth}  
  \centering
  \centerline{\includegraphics[width=4cm]{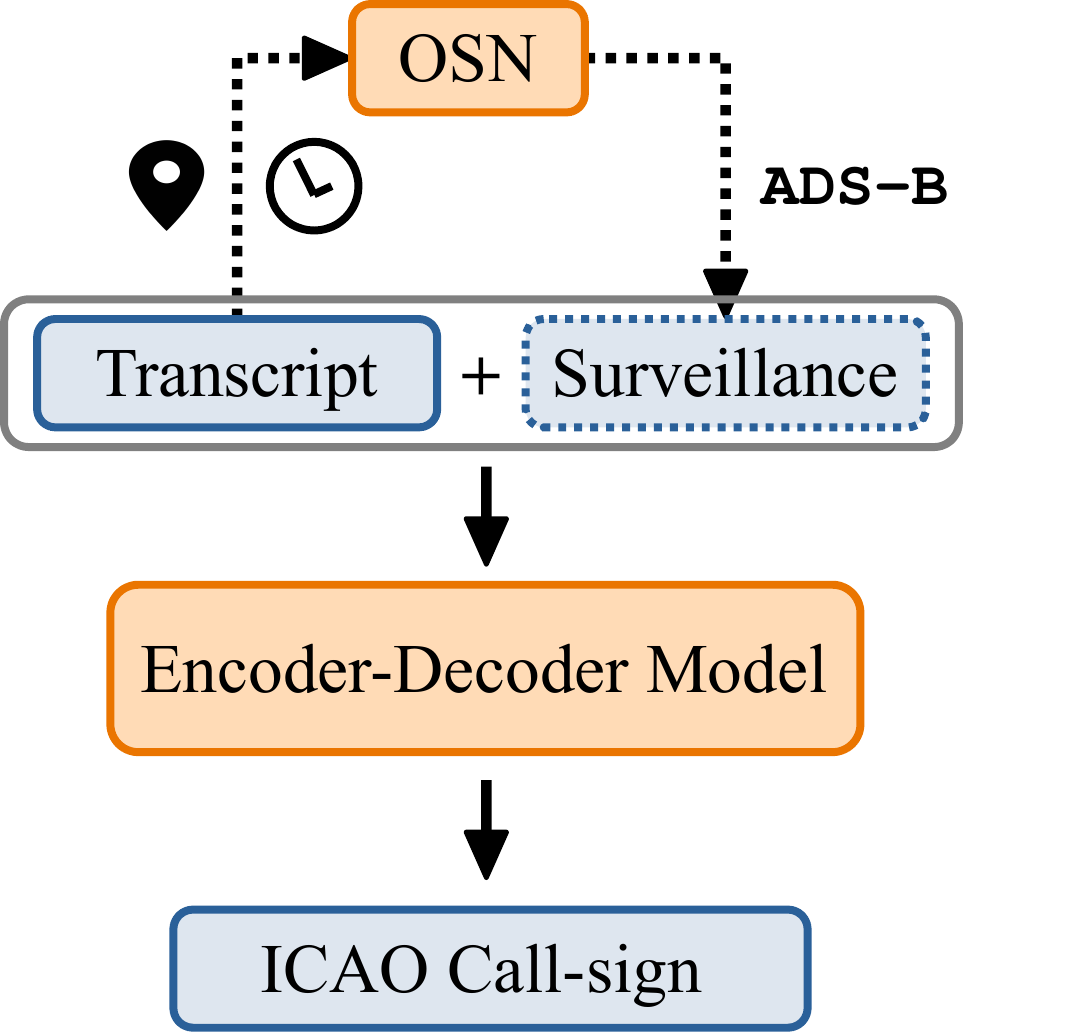}}
\end{minipage}
\caption{The CRU system. The dotted path marks the optional surveillance retrieval via OSN with the aid of the transcripts timestamp and VHF receiver location.}
\label{fig:model}
\end{figure}

Context integration is necessary, since not all of the information loss through the ASR system can be recovered. If for example \texttt{five} would be missing in \texttt{Ryanair eight \textbf{five} three kilo}, the remaining \texttt{Ryanair eight three kilo} would be the wrong, but valid call-sign. A conventional CRU system would therefore predict  \texttt{RYR83K} as target call-sign instead of \texttt{RYR853K}. Adding surveillance call-signs as additional input, as shown in \autoref{fig:model}, allows the model to compensate the missing information in the transcript. With the timestamp of the transcript and its recording location, surveillance information can be retrieved from OSN (dotted path in \autoref{fig:model}), like described in \autoref{sec:data}.

\section{Experimental Setup}\label{sec: experimental setup}
The basis of our CRU model is the \texttt{EncoderDecoderModel} from the Hugging Face transformers library \cite{Wolf2020}, which showed superior performance over other designs in prior experiments. The language model head of this architecture allows also to predict call-signs without surveillance information. For both, encoder and decoder the pretrained \texttt{bert-base-uncased} model is used, to make use of the beneficial effect of using pretrained models for sequence-to-sequence tasks \cite{Rothe2020}.
 \begin{table}[ht]
\centering
\caption{Accuracy on the LiveATC testsets. The call-sign recognition models are trained on the augmented Airbus dataset with different WERs. Underlined accuracy scores symbolize the best vanilla recognition model, while bold scores mark the best model overall.}
\setlength{\tabcolsep}{4pt}
\begin{tabular}[t]{l|cccccccc}
\toprule
   & \multicolumn{8}{c}{Accuracy on LiveATC test sets}\\
Taining &  \multicolumn{2}{c|}{WER 0} & \multicolumn{2}{c|}{WER 28.4} & \multicolumn{2}{c|}{WER 28.9} & \multicolumn{2}{c}{WER 33.1 } \\
~~~sets &  \multicolumn{1}{c}{Van} & \multicolumn{1}{c|}{Sur} &  \multicolumn{1}{c}{Van} & \multicolumn{1}{c|}{Sur}  &  \multicolumn{1}{c}{Van} & \multicolumn{1}{c|}{Sur}  &  \multicolumn{1}{c}{Van} & \multicolumn{1}{c}{Sur}  \\
\midrule
WER 0 &  39.8 & \textbf{89.4} & 31.0 & 74.0 & 27.8 & 70.3 & 11.2 & 45.2 \\
WER 7 &  \underline{60.2} & 88.6 & \underline{47.2} & \textbf{78.4} & \underline{44.0} & \textbf{73.3} & \underline{16.4} & \textbf{57.0} \\
WER 30&  56.0 & 85.6 & 46.6 & 73.6 & 42.8 & 68.5 & 15.4 & 47.4 \\
\bottomrule
\end{tabular}
\label{tab:LiveATC}
\end{table}%

 \begin{table}[ht]
\centering
\caption{Accuracy on the Malorca testsets. The call-sign recognition models are trained on the augmented Airbus dataset with different WERs. Underlined accuracy scores symbolize the best vanilla recognition model, while bold scores mark the best model overall.}
\setlength{\tabcolsep}{4pt}
\begin{tabular}[t]{l|cccccccc} 
\toprule
   & \multicolumn{8}{c}{Accuracy on Malorca test sets}\\
Taining &  \multicolumn{2}{c|}{WER 0} & \multicolumn{2}{c|}{WER 6.42} & \multicolumn{2}{c|}{WER 7.27} & \multicolumn{2}{c}{WER 8.47 } \\
~~~sets &  \multicolumn{1}{c}{Van} & \multicolumn{1}{c|}{Sur} &  \multicolumn{1}{c}{Van} & \multicolumn{1}{c|}{Sur}  &  \multicolumn{1}{c}{Van} & \multicolumn{1}{c|}{Sur}  &  \multicolumn{1}{c}{Van} & \multicolumn{1}{c}{Sur}  \\
\midrule
WER 0 &  49.5 & 85.6 & 50.6 & 82.4 & 47.4 & 79.5 & 44.2 & 75.6 \\
WER 7 &  53.8 & \textbf{87.5} & 53.6 & 84.9 & 50.4 & 83.5 & 46.8 & 80.7 \\
WER 30&  \underline{54.8} & 87.3 & \underline{54.7} & \textbf{85.0} & \underline{50.9} & \textbf{83.7} & \underline{47.2} & \textbf{81.0} \\
\bottomrule
\end{tabular}
\label{tab:Malorca}
\end{table}%
 Since ATC speech transcripts differ highly from standard text, a domain adaptation is done by pretraining BERT (\texttt{bert-base-uncased}) on ATC transcripts using masked language modeling. The training of the CRU models is done on the augmented Airbus datasets. For each augmented dataset listed in \autoref{tab:datasets} (WER 0, WER 7 and WER 30) a split of 40k/10k/10k for train/val/test sets is used. The models are either trained with (Sur) or without (Van) surveillance information. The transcript and the surveillance call-signs are concatenated and embedded into a single vector. This single or cross-encoder design allows interactions between the transcript and context from lower layers of the model on. The overall architecture for the Sur and Van models is the same, to ensure a fair comparison. The trained models are tested on the LiveATC and Malorca test sets listed in \autoref{tab:datasets}. The performance of all models is measured as accuracy or call-sign accuracy (CSA). 

\section{Experimental Results}
\subsection{Surveillance Incorporation}

\begin{figure*}[ht] 
\hspace{0.4cm}
\begin{minipage}[b]{0.3\linewidth}  
  \centering
  \centerline{\includegraphics[width=5cm]{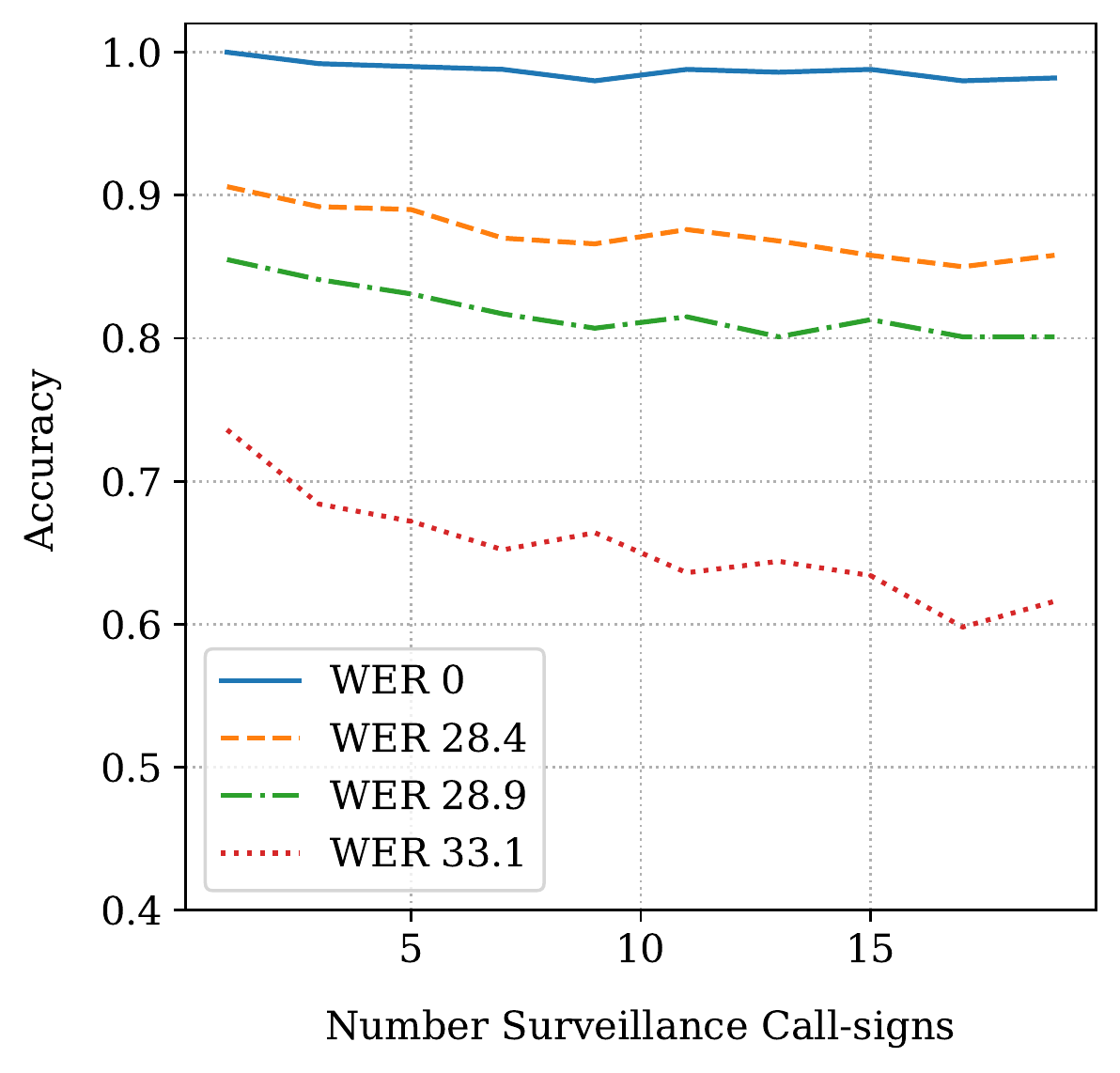}}
\end{minipage}
%
\begin{minipage}[b]{0.3\linewidth}
  \centering
  \centerline{\includegraphics[width=5cm]{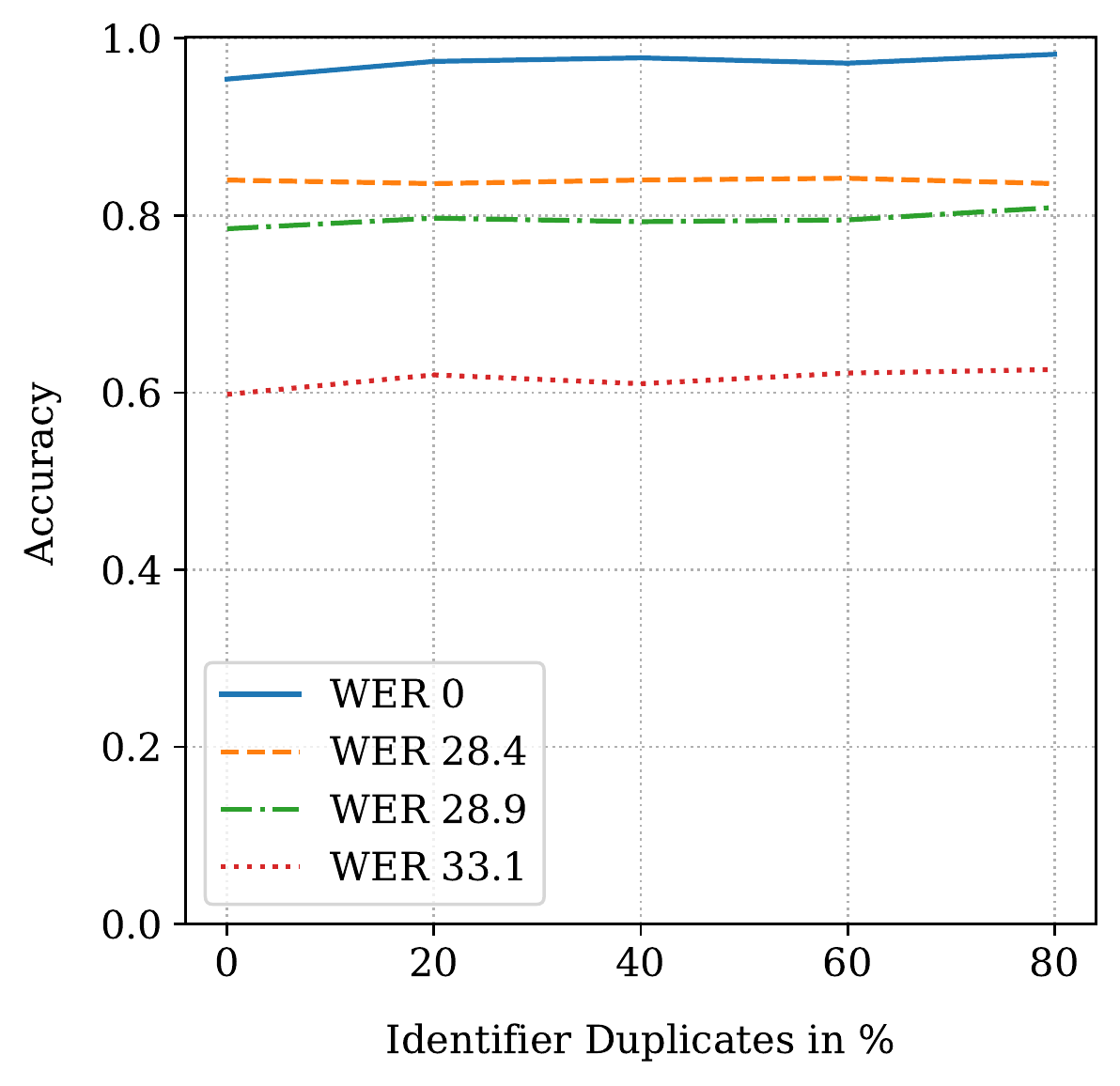}}
\end{minipage}
%
\begin{minipage}[b]{0.3\linewidth}  
  \centering
  \centerline{\includegraphics[width=5cm]{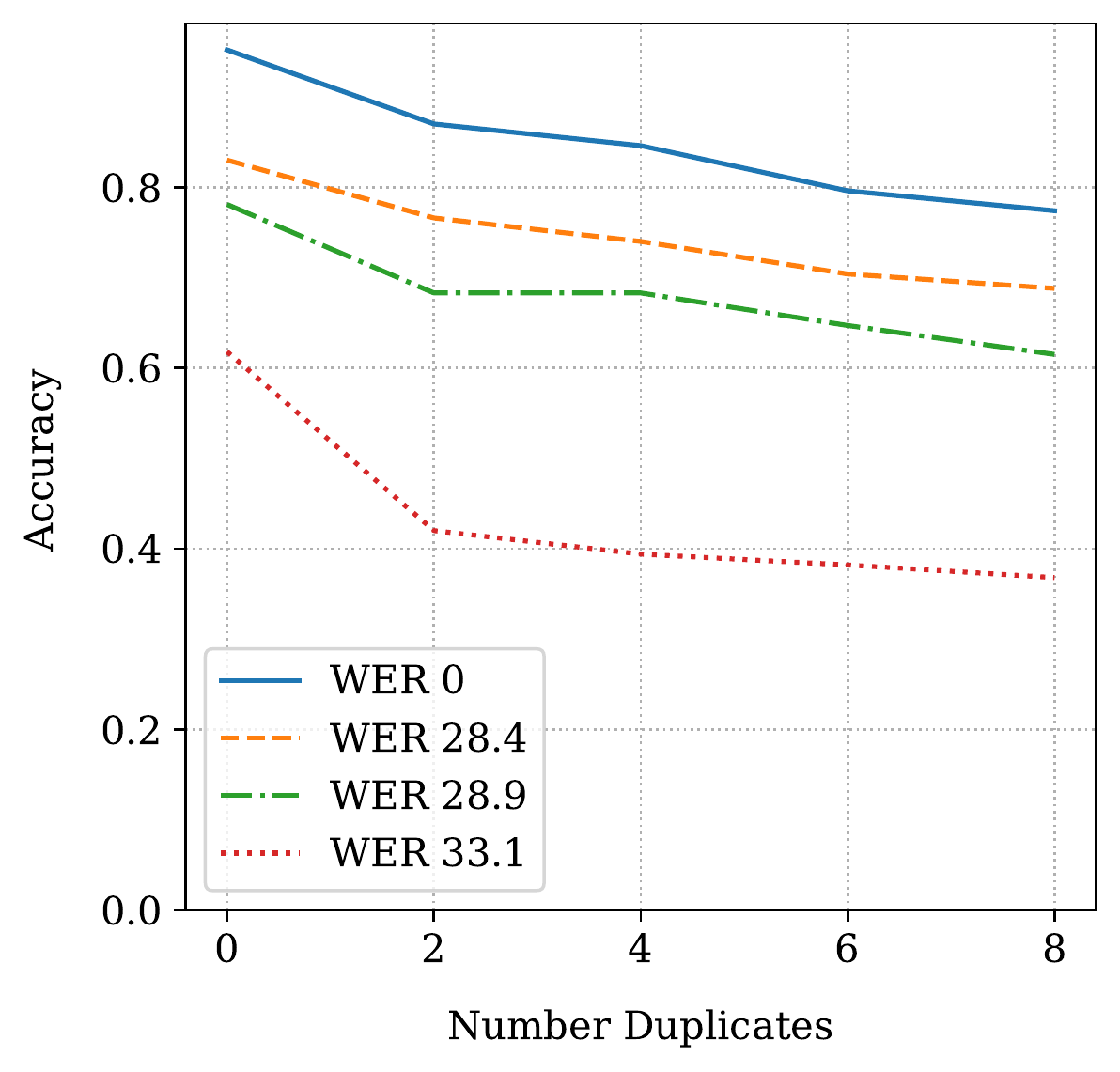}}
\end{minipage}
\caption{Change of accuracy depending on (left) the number of call-signs in the surveillance data; (middle) the relative number of additional call-signs in the surveillance information containing the same call-sign identifier as the target call-sign; (right) the number of additional call-signs in the surveillance information containing the same call-sign number as the the target call-sign.}
\label{fig:identifierDuplicatesres}
\end{figure*}

Feeding the model surveillance call-signs not only allows recovering noisy ASR transcripts, that are lacking e.g. the airline identifier. The surveillance allows the model also to predict call-signs containing airline identifiers, that did not appear in the training data. Additionally the surveillance call-signs decrease the target space for the model. \par
\autoref{tab:LiveATC} and \autoref{tab:Malorca} show the comparison between CRU models incorporating surveillance (Sur) and not incorporating surveillance (Van). The models that include surveillance call-signs outperform the vanilla models on every test set. On the high noise transcripts of the LiveATC dataset (WER 33.1), the benefit of the additional information shows the best. The vanilla network is here outperformed by a factor of 3-4. As an example of the recovering capabilities for noisy call-signs, we are able to predict 57\% of the ICAO call-signs from the LiveATC transcripts (WER 33.1). Although they contain only 27\% correct call-signs. This means an increase of 30\%. 

\subsection{Noisy Training Data}

To give the models more robustness against ASR noise, they are trained on different WER-level training data. On the Malorca data, the models trained on noisy transcripts (WER 7 and WER 30) outperform the model trained on clean transcripts (WER 0) on every test set. Both, the surveillance and the vanilla models benefit on similar levels from the training on noisy data, while the highest performance boost is reached on the noisiest test set (WER 8.47) from 75.6\% accuracy to 81.0\%. On the noisier LiveATC test sets the overall mean accuracy of the vanilla model trained on WER 7 data is around 1.5 times higher than the accuracy of the model trained on noise-free data. Raising the WER of the training data further from 7 to 30 leads only to a small improvement on the high WER Malorca test sets.\par
The results show the benefit of training the model on (simulated) noisy transcripts if the target input of the model is ASR recognizer output. But more importantly, they also show, that even if there is just clean data available for training, including the surveillance call-signs is a necessary condition to reach maximum performance.

\subsection{Surveillance Fluctuation Robustness}\label{sec:robustness}

We investigate the robustness of our model against the three main surveillance parameters: number of call-signs in the surveillance, number of airline identifier duplicates and number of call-sign number duplicates. The evaluations are done on the LiveATC dataset by altering the surveillance information. The model trained on the WER 7 dataset is used for these tests since it performs the best on the noisy LiveATC test sets.\par
A higher number of surveillance call-signs increases the search space for the model. By increasing the surveillance size from 1 to 19, the accuracy decreases by 5\% on the WER 28.4 test data, while there is a decrease of 12\% on the WER 33.1 test set as \autoref{fig:identifierDuplicatesres} shows. Intuitively this is clear since on noisy data, the model has to rely more on the additional context information.\par
Several airplanes of the same airline can be in the same airspace resulting in call-signs with an identical identifier (e.g \texttt{DLH124, DLH9M, DLH69F}). For the LiveATC and Malorca test set, each identifier in the surveillance occurs 1.45 respectively 1.9 times. \autoref{fig:identifierDuplicatesres} shows, that the recognizer is very robust against airline identifier duplicates. Even with 80\% of the surveillance call-sign identifiers being identical to the target identifier call-sign, there is no drop in accuracy.\par
In contrast to identifier duplicates, having the same call-sign number in the surveillance information (e.g. \texttt{DLH83K, CSA83K, RYR83K}) is quite rare. In the LiveATC dataset, only in  2.7\% of the cases, a call-sign number appears twice in the surveillance information. With one duplicate of the target call-sign, which is already higher then what can be expected in the real-life scenario, the accuracy drop on the WER 28.4 and WER 28.9 dataset stays below 5\% as \autoref{fig:identifierDuplicatesres} shows. For the high noise dataset, the drop is around 10\%.

\section{Conclusion}
In this work, we have introduced a method for enhancing call-sign recognition and understanding (CRU) by incorporating context information in the form of surveillance call-signs without changing the model architecture. We have shown that this improves the call-sign accuracy up to 4 times. Our data augmentation pipeline allows to generate training data for specific airspaces, even if there are no transcripts available for that region. We have shown that introducing ASR noise in the data augmentation pipeline improves the vanilla model performance up to 1.5 times.\par
We can show that our models are robust against the occurrence of multiple surveillance call-signs containing the same identifier. The number of included surveillance call-signs included should be kept as low as possible since the call-sign accuracy decreases linearly with the number of the surveillance call-signs. For the rare case of an additional call-sign occurring with the target call-sign number, we can show that the accuracy drop stays below 5\% for the low call-sign WER test sets and under 10\% for the high WER call-sign test set.\par
In the future, we want to look also at other context incorporation methods. We also plan to adapt our model to other named entities appearing in ATC transcripts like commands and values.

\label{sec:ref}

\bibliographystyle{IEEEbib}
\bibliography{strings,ASRU}

\begin{thebibliography}{10}

\bibitem{Eskilsson2020}
S.~Eskilsson, H.~Gustafsson, S.~Khan, and A.~Gurtov,
\newblock ``{Demonstrating ADS-B and CPDLC Attacks with Software-Defined
  Radio},''
\newblock {\em Integrated Communications, Navigation and Surveillance
  Conference, ICNS}, vol. 2020-Septe, pp. 1--9, sep 2020.

\bibitem{Kleinert2018}
M.~Kleinert, H.~Helmke, G.~Siol, H.~Ehr, A.~Cerna, C.~Kern, D.~Klakow,
  P.~Motlicek, Y.~Oualil, M.~Singh, and A.~Srinivasamurthy,
\newblock ``{Semi-supervised adaptation of assistant based speech recognition
  models for different approach areas},''
\newblock in {\em AIAA/IEEE Digital Avionics Systems Conference - Proceedings}.
  dec 2018, vol. 2018-Septe, Institute of Electrical and Electronics Engineers
  Inc.

\bibitem{Srinivasamurthy2018a}
A.~Srinivasamurthy, P.~Motlicek, M.~Singh, Y.~Oualil, M.~Kleinert, H.~Ehr, and
  H.~Helmke,
\newblock ``{Iterative learning of speech recognition models for air traffic
  control},''
\newblock {\em Proceedings of the Annual Conference of the International Speech
  Communication Association, INTERSPEECH}, vol. 2018-Septe, no. September, pp.
  3519--3523, 2018.

\bibitem{Zuluaga-Gomez2020b}
J.~Zuluaga-Gomez, P.~Motlicek, Q.~Zhan, K.~Vesely, and R.~Braun,
\newblock ``{Automatic speech recognition benchmark for air-traffic
  communications},''
\newblock in {\em Proceedings of the Annual Conference of the International
  Speech Communication Association, INTERSPEECH}, jun 2020, vol. 2020-Octob,
  pp. 2297--2301.

\bibitem{Shore}
T.~Shore, F.~Faubel, H.~Helmke, and D.~Klakow,
\newblock ``{Knowledge-based word lattice rescoring in a dynamic context},''
\newblock in {\em 13th Annual Conference of the International Speech
  Communication Association 2012, INTERSPEECH 2012}, 2012, vol.~2, pp.
  1082--1085.

\bibitem{Schmidt}
A.~Schmidt, Y.~Oualil, O.~Ohneiser, M.~Kleinert, M.~Schulder, A.~Khan,
  H.~Helmke, and D.~Klakow,
\newblock ``{Context-based recognition network adaptation for improving on-line
  ASR in air traffic control},''
\newblock in {\em 2014 IEEE Workshop on Spoken Language Technology, SLT 2014 -
  Proceedings}, 2014, pp. 13--15.

\bibitem{Oualil}
Y.~Oualil, M.~Schulder, H.~Helmke, A.~Schmidt, and D.~Klakow,
\newblock ``{Real-time integration of dynamic context information for improving
  automatic speech recognition},''
\newblock in {\em Proceedings of the Annual Conference of the International
  Speech Communication Association, INTERSPEECH}, 2015, vol. 2015-Janua, pp.
  2107--2111.

\bibitem{KocourMartin2019}
M.~Kocour, K.~Veselý, A.~Blatt, J.~Zuluaga Gomez, I.~Szöke, J.~Černocký,
  D.~Klakow, and P.~Motlicek,
\newblock ``{Boosting of Contextual Information in ASR for Air-Traffic
  Call-Sign Recognition},''
\newblock pp. 3301--3305, 2021.

\bibitem{Nigmatulina2021}
I.~Nigmatulina, R.~Braun, J.~Zuluaga-Gomez, and P.~Motlicek,
\newblock ``{Improving callsign recognition with air-surveillance data in
  air-traffic communication},'' aug 2021.

\bibitem{Povey2011}
D.~Povey, G.~Boulianne, L.~Burget, P.~Motlicek, and P.~Schwarz,
\newblock ``{The Kaldi Speech Recognition},''
\newblock {\em IEEE 2011 Workshop on Automatic Speech Recognition and
  Understanding}, , no. January, 2011.

\bibitem{Pellegrini2018}
T.~Pellegrini, J.~Farinas, E.~Delpech, and F.~Lancelot,
\newblock ``{The airbus air traffic control speech recognition 2018 challenge:
  Towards ATC automatic transcription and call sign detection},''
\newblock in {\em Proceedings of the Annual Conference of the International
  Speech Communication Association, INTERSPEECH}, oct 2019, vol. 2019-Septe,
  pp. 2993--2997.

\bibitem{Gupta2019a}
V.~Gupta, L.~Rebout, G.~Boulianne, P.~M{\'{e}}nard, and J.~Alam,
\newblock ``{CRIM's Speech Transcription and Call Sign Detection System for the
  ATC Airbus Challenge Task},''
\newblock in {\em Interspeech 2019}. sep 2019, pp. 3018--3022, International
  Speech Communication Association.

\bibitem{Devlin2019}
J.~Devlin, M.~Chang, K.~Lee, and K.~Toutanova,
\newblock ``{BERT: Pre-training of deep bidirectional transformers for language
  understanding},''
\newblock in {\em NAACL HLT 2019 - 2019 Conference of the North American
  Chapter of the Association for Computational Linguistics: Human Language
  Technologies - Proceedings of the Conference}, 2019, vol.~1, pp. 4171--4186.

\bibitem{Delpech2019}
E.~Delpech, M.~Laignelet, C.~Pimm, C.~Raynal, M.~Trzos, A.~Arnold, and
  D.~Pronto,
\newblock ``{A real-life, French-accented corpus of air traffic control
  communications},''
\newblock in {\em LREC 2018 - 11th International Conference on Language
  Resources and Evaluation}, 2019, pp. 2866--2870.

\bibitem{Schafer2014}
M.~Sch{\"{a}}fer, M.~Strohmeier, V.~Lenders, I~Martinovic, and M.~Wilhelm,
\newblock ``{Bringing up OpenSky: A large-scale ADS-B sensor network for
  research},''
\newblock in {\em IPSN 2014 - Proceedings of the 13th International Symposium
  on Information Processing in Sensor Networks (Part of CPS Week)}. 2014, pp.
  83--94, IEEE Computer Society.

\bibitem{Wolf2020}
T.~Wolf, L.~Debut, V.~Sanh, J.~Chaumond, C.~Delangue, A.~Moi, P.~Cistac,
  T.~Rault, R.~Louf, M.~Funtowicz, J.~Davison, S.~Shleifer, P.~von Platen,
  C.~Ma, Y.~Jernite, J.~Plu, and C~Xu,
\newblock ``{Transformers: State-of-the-Art Natural Language Processing},''
\newblock 2020, pp. 38--45.

\bibitem{Rothe2020}
S.~Rothe, S.~Narayan, and A.~Severyn,
\newblock ``{Leveraging Pre-trained Checkpoints for Sequence Generation
  Tasks},''
\newblock Tech. {R}ep., 2020.

\end{thebibliography}

\end{document}